# An Improved Convolutional Neural Network System for Automatically Detecting Rebar in GPR Data

**Zhongming Xiang, Ph.D. Candidate[1], Abbas Rashidi, Assistant Professor[2], and Ge (Gaby) Ou, Assistant Professor [3]**

[1]Dept. of Civil and Environmental Engineering, Univ. of Utah, Salt Lake City, UT 84112; e-mail: zhongming.xiang@utah.edu
[2]Dept. of Civil and Environmental Engineering, Univ. of Utah, Salt Lake City, UT 84112; e-mail: abbas.rashidi@utah.edu
[3]Dept. of Civil and Environmental Engineering, Univ. of Utah, Salt Lake City, UT 84112; e-mail: Ge.Ou@utah.edu

**ABSTRACT**

As a mature technology, Ground Penetration Radar (GPR) is now widely employed in detecting rebar and other embedded elements in concrete structures. Manually recognizing rebar from GPR data is a time-consuming and error-prone procedure. Although there are several approaches to automatically detect rebar, it is still challenging to find a high resolution and efficient method for different rebar arrangements, especially for closely spaced rebar meshes. As an improved Convolution Neural Network (CNN), AlexNet shows superiority over traditional methods in image recognition domain. Thus, this paper introduces AlexNet as an alternative solution for automatically detecting rebar within GPR data. In order to show the efficiency of the proposed approach, a traditional CNN is built as the comparative option. Moreover, this research evaluates the impacts of different rebar arrangements, and different window sizes on the accuracy of results. The results revealed that: (1) AlexNet outperforms the traditional CNN approach, and its superiority is more notable when the rebar meshes are densely distributed; (2) the detection accuracy significantly varies with changing the size of splitting window, and a proper window should contain enough information about rebar; (3) uniformly and sparsely distributed rebar meshes are more recognizable than densely or unevenly distributed items, due to lower chances of signal interferences.

## INTRODUCTION

Extracting necessary information about the number, location(s), and size(s) of embedded rebar at existing concrete elements is a major task for civil engineers. As a popular Non-Destructive Testing (NDT) method, Ground Penetration Radar (GPR) is capable of detecting rebar and other embedded metallic without causing any destructions to concrete elements. The technology has been proved to be very efficient in various projects and under different settings (Kaur et al. 2016; Eisenmann et al. 2017). Based on the propagation principle of electromagnetic waves (EM), rebar is presented as hyperbolic signatures in GPR data. As a result, extracting necessary information and interpreting the hyperbolic signatures is a critical step toward automated detection of rebar. Several approaches have been suggested by researchers and practitioners to handle this job (Dou et al. 2017; Lee and Mokji, 2014; Yuan et al. 2018). One of the most effective techniques is implementing machine learning algorithms, such as convolution neural network (CNN), support vector machines (SVM), BP neural network, etc.

Since image is the most intuitive form of GPR data, CNN, a powerful tool for several image processing algorithms (Chua and Roska, 1993), is very suitable for interpreting GPR data. One major advantage is that CNN does not require extracting any features from the raw data, which eliminates the need for some extra computing steps necessary for other machine learning methods (Guyon and Elisseeff, 2006). Several studies have been conducted on applications of CNN for interpreting GPR data. CNN was first implemented by Besaw and Stimac (2015) to interpret GPR data of buried explosive hazards. The results illustrated that the accuracy could increase by 10% compared to traditional feature extraction approaches. Similarly, Lameri et al. (2017) employed CNN to detect buried landmines, and their study showed that the accuracy could raise up to 95% on real GPR data with minimal pre-processing procedures. For recognizing rebar, Dinh et al. (2018) used CNN to locate and detect rebar in bridge decks. Necessary image processing methods were applied to obtain high quality GPR data for CNN, and the reported accuracy level was higher than 95.75%. One characteristic of those studies is that the buried rebar meshes were not densely distributed. In other words, low signal interface occurred in GPR data. However, in most exiting concrete elements (e.g. column, shear wall, slab, etc.), the distribution of rebar is quite dense, and the reflected hyperbolas in GPR patterns are often rambling. As a result, a more efficient method is required to deal with densely distributed rebar meshes at concrete elements.

As an improved version of CNN, AlexNet showed more promising results in image recognition applications through the ImageNet competition in 2012 (Russakovsky et al. 2015). This technique has been widely used for recognizing targets and shown distinctly high levels of accuracy. Due to implementing deeper layers and exitance of several new features in the network, a trained AlexNet is very robust. It can overcome the problem of scattered distribution of dense rebar meshes. For the first time, this paper applies the AlexNet to detect the existence of rebar in reinforced concreted elements. In order to illustrate the superiority of AlexNet, a traditional CNN has been built for comparing detection accuracy. This study also evaluates the impacts of window sizes (used for dividing the entire GPR image into training and testing segments) on accuracy of final results. The proposed method has been tested on three major structural elements: concrete columns, shear walls, and concrete slabs. The following sections describe necessary steps for constructing a CNN AlexNet system for detecting rebar, as well as necessary experimental settings to evaluate the proposed system and obtained results.

**RESEARCH METHODOLOGY**

The ultimate goal of this research is to propose a more efficient method for automatically detecting rebar in concrete elements. To achieve this goal, a novel CNN AlexNet has been constructed. In parallel, a traditional CNN has been built for comparison purposes. architecture details of the architecture of the two deep networks, as well as the necessary pre-processing steps to implement these systems are described here:

*AlexNet Architecture*

An AlexNet system consists of 8 layers, including 5 convolutional and 3 full-connected layers (Krizhevsky et al. 2012). Figure 1 depicts the network structure of AlexNet. As the uniform input data of the network, the image size is 227×227×3. In the layer Conv1, 96 convolution kernels are set to process the input data. Meanwhile, the activation function ReLU is employed to ensure that values in the feature map are in a reasonable range. A max-pooling layer and the local response normalization (LRN) are used as well in Conv1. The layers 2-5 resemble Conv1 except that Conv3 and Conv4 do not contain pooling layer. After processed by convolution layers, a new feature map is sent to the full-connected layers 6-8. Similar to a traditional CNN, AlexNet considers the weights as the connections of the neurons between different layers. In order to reduce overfitting, half of neurons are randomly dropped out. In addition, the Softmax is employed to classify the categories in a reasonable way. The final classification is conducted based on the highest possibilities of input data to belong to any desired categories.

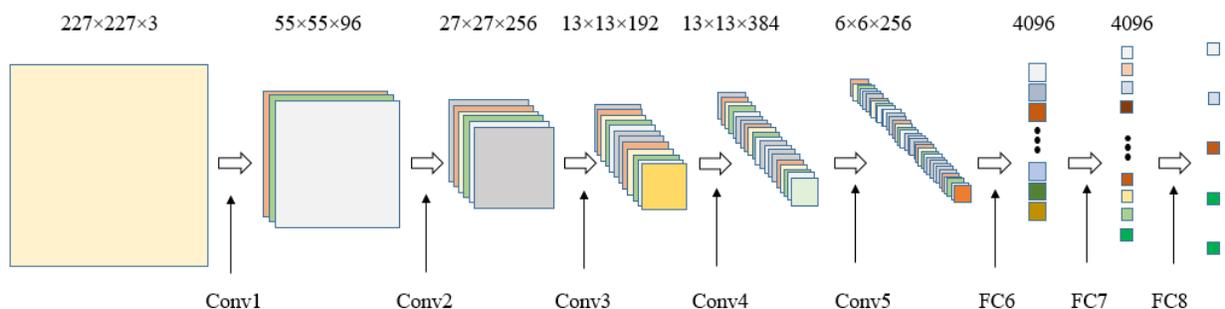

**Figure 1. The network structure of AlexNet**

As a deep learning method, AlexNet incorporates several new features into CNN. These additions improve both recognition accuracy and computation efficiency (Krizhevsky et al. 2012). Different features of AlexNet, compared to traditional CNN, are summarized in Table 1.

**Table 1. Improvement details of AlexNet compared to traditional CNN**

| Items | Traditional CNN | AlexNet | Improvement |
|---|---|---|---|
| Activation function | Sigmoid | ReLU | Avoid the gradient diffusion in deep network |
| Neurons in use | All of the neurons | A portion of neurons | Reduce overfitting |
| Pooling layer | Average-pooling | Max-pooling | Retain the significant features |
| Neuron activity | - | LRN | Improve the generalization |
| Operation mode | CPU | GPU | Reduce the computation time |
| Data size | Original data | Data Augmentation | Reducing overfitting |

*Comparative CNN*

As the next step, and in order to compare the performance of AlexNet with traditional CNN, another neural network named TraNet, is constructed. TraNet contains 6 layers, in which layers 1, 3, and 5 are built as convolutional layers, layer 2 and 4 are pooling layers, and layer 6 is both the full-connected layer and the output layer. Detailed settings of TraNet are summarized in **Table 2**. For comparison purposes, and to achieve more uniform results, some parameters such as learning rate, maximum epochs, and weight learn rate factor have been set similarly in both TraNet and AlexNet systems.

Table 2. Major parameters of the constructed traditional CNN system (TraNet)

| Layers | Parameters |
|---|---|
| Convolution 1 | 8@3×3; Stride: 1; Padding: 0; RELU; Bath Normalization |
| Pooling 2 | 2×2; Stride: 2 |
| Convolution 3 | 16@3×3; Stride: 1; Padding: 0; RELU; Bath Normalization |
| Pooling 4 | 2×2; Stride: 2 |
| Convolution 5 | 32@3×3; Stride: 1; Padding: 0; RELU; Bath Normalization |
| Full Connected 6 | Output neuron numbers: 4; Activation function: Softmax |

*GPR Data Pre-processing*

Normally, a typical GPR data contains several hyperbolic signatures reflected by existence of rebar. In order to make clear classification, a rectangular window has been applied to split the whole GPR image into several smaller parts. The main assumption for setting theses windows is that no more than one complete hyperbolic signature should exist in each part. As presented in Figure 2, there are four typical shapes in the segmented images: left, peak, right, and other. If the shape in the sub-image is notably left, peak or right, it will be correspondingly classified as 'left', 'peak' or 'right'. Otherwise, and if the segment image does not include any recognizable shape, it will be classified as 'other'. Meanwhile, based on the requirement of input data, all of these images are resized to 227×227×3 prior to using for training or testing the proposed network.

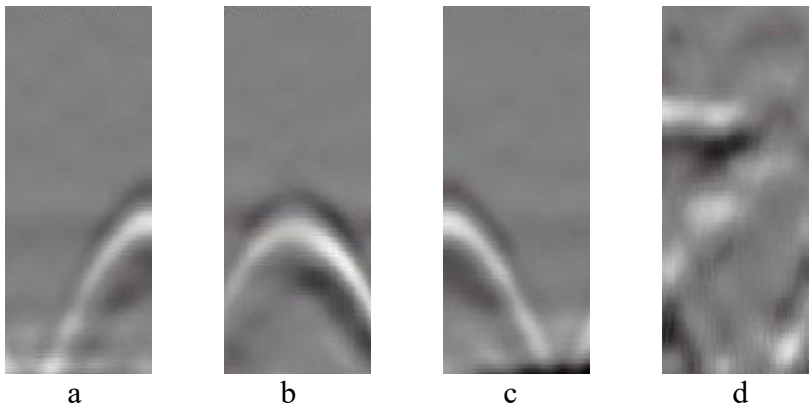

a       b       c       d
Figure 2. The four typical types of input data: a-left, b-peak, c-right, and d-other

During the training step, it is found that the size of the dividing window has a significant impact on accuracy of detection results. To further evaluate this impact, four different windows sizes have been selected: 120×30, 150×50, 200×80, and 250×100 (pixels). The visual effects of different window sizes are demonstrated in Figure 3.

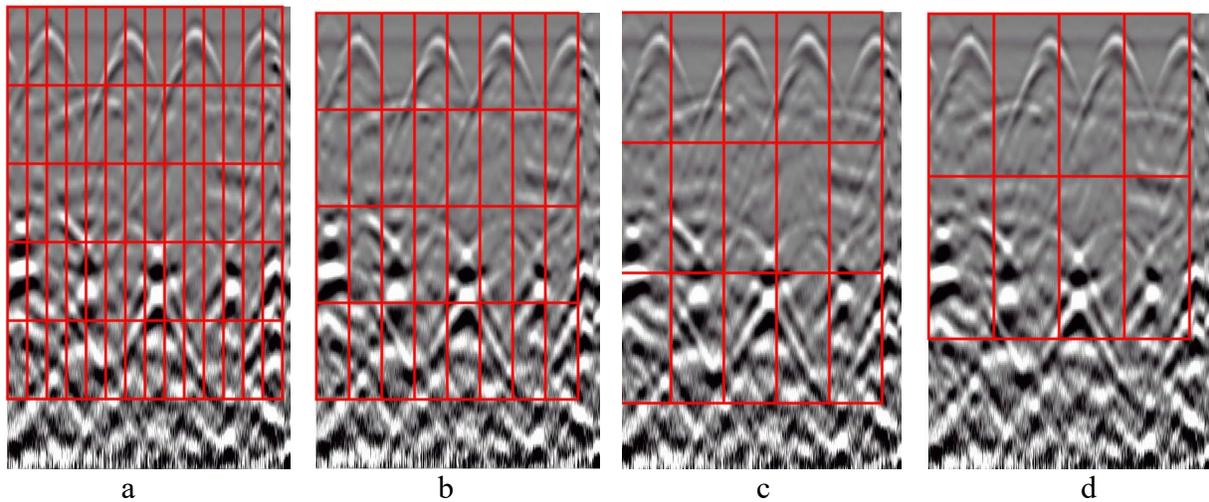
**Figure 3. Examples of different window sizes: a-120×30; b-150×50; b-200×80; d-250×100**

**EXPERIMENTAL SETUP AND CASE STUDY**

For demonstrating the efficiency of the proposed AlexNet system, a number of experiments have been conducted. This section briefly reviews the selected experimental setup as well as the obtained results.

*Experimental Data*

Several reinforced concrete elements in a newly renovated building have been selected as testbed for this study. Three major building elements have been used as the case studies: one concrete column with the size of 305 mm×305 mm, one concrete shear wall with thickness of 305 mm, and a 203 mm thick concrete slab. Figure 4 depicts the three selected elements and the corresponding scanning directions. The stirrup rebar, horizontal rebar, and rebar placed in two directions are chosen as objects of interest in the column, the shear wall, and the slab, respectively.

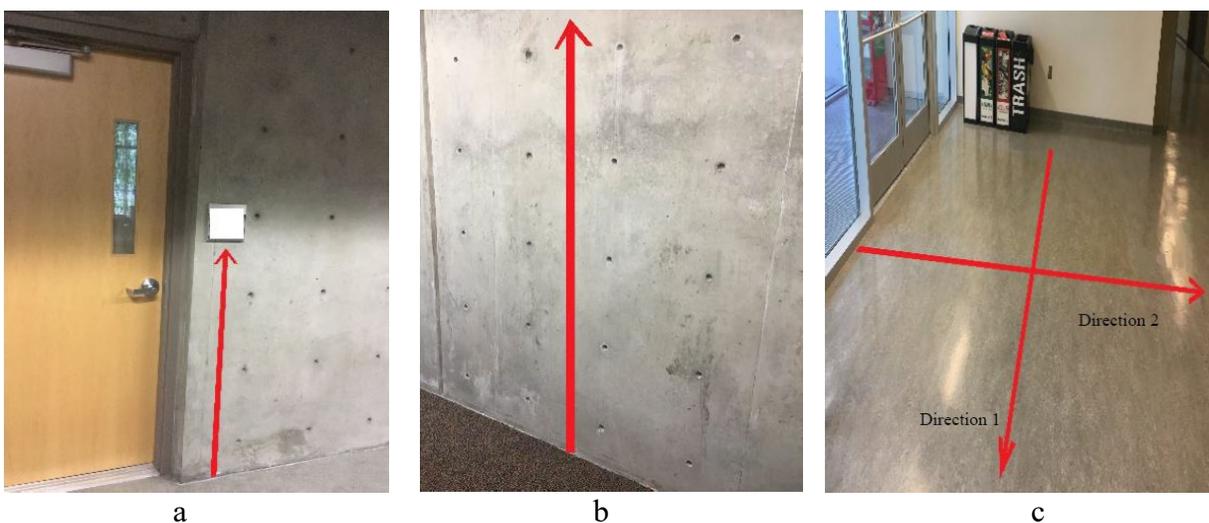
**Figure 4. Three Case studies used as testbed: a-Column; b-Shear Wall; c-Suspended Slab**

The popular all-in-one ground penetrating system, StructureScan Mini XT, with the central frequency 2.7 GHz, has been implemented for scanning the case studies and generating raw data in the form of 2D images. Figure 5 shows the typical GPR image generated by this machine. In this project, 48 images have been collected. All these images have been divided into relatively small parts for training purposes. 80% of these small parts have used as training data, and the rest were considered as test data.

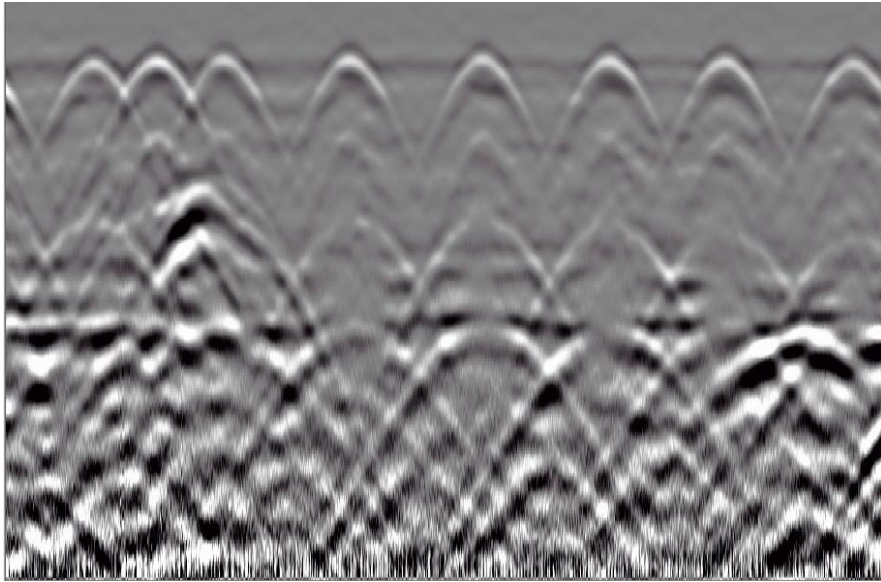

**Figure 5. A typical GPR imagery data obtained by scanning a sample concrete element**

*Results and Discussion*

The purpose of using the trained AlexNet and TraNet systems is to predict type of label in each small part. Table 3 summarizes the testing accuracy achieved by using two networks with different window sizes. As shown in the table, regardless of the window size, AlexNet is capable of generating more accurate results. For certain window sizes (e.g. 120×30, 150×50, and 250×100), the accuracy of AlexNet is at least 8% higher than that of TraNet.

**Table 3. Testing accuracy of AlexNet and TraNet**

| Window Size | TraNet | | | AlexNet |
|---|---|---|---|---|
| | Size of input images | | | |
| | 28×28×1 | 32×32×1 | 227×227×1 | 227×227×1 |
| 120×30 | 61.27% | 57.82% | 33.16% | 72.68% |
| 150×50 | 79.19% | 78.73% | 52.49% | 87.78% |
| 200×80 | 91.21% | 91.31% | 76.92% | 94.51% |
| 250×100 | 74.62% | 77.69% | 73.85% | 82.31% |

When it comes to different window sizes, the accuracy is highest in the case of 200×80 window size. This phenomenon is mainly due to the fact that this specific window size contains diverse parts of the hyperbolas. Considering the cases shown in Figure 3-a and b, if the selected window is too small, it may not contain enough information about the hyperbola. On the other hand, if the selected window is too large, it may contain the information of more than one rebar (Figure 3-d).

The accuracy levels for the three elements are separately plotted in Figure 6. In order to make the comparison concise and clear, we only show one case of TraNet where the size of the input image is 28×28×1. Through analyzing the GPR data of the three elements, it is found that the rebar distribution in column is largest and most even due to the small signal interference from the neighboring rebar. On the contrary, there exist a lot of signal interference in the GPR data of the slab. As a result, the accuracy rate is highest for the column cases and lowest for the slab cases (Figure 6).

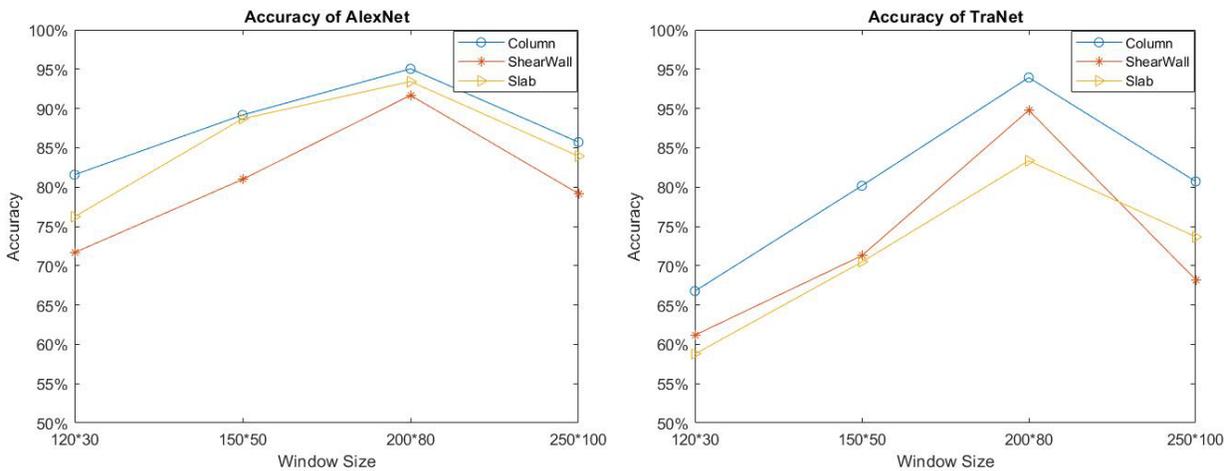

**Figure 6. Accuracy of implementing AlexNet and TraNet systems with different window**

**CONCLUSION AND FUTURE WORK**

This paper provided a novel approach of recognizing rebar in GPR patterns with one notable convolutional neural network AlexNet. Based on the comprehensive analysis of the network architecture, AlexNet was trained and tested by using GPR data scanned from three concrete elements (one column, one shear wall, and one suspended slab). The comparisons of detection accuracy between AlexNet and traditional CNN has been conducted. Meanwhile, for analyzing the size influence of the windows on rebar detection, four different window sizes have been set and compared. At the end, the accuracy among the three elements were discussed as well. In summary, the following conclusions have been learned:
- Compared with traditional CNN, AlexNet could achieve higher levels of accuracy in recognizing the rebar in actual constructed facilities.
- Variations in sizes of splitting window could remarkably affect the recognition result. This situation is more crucial for traditional CNN, and AlexNet is more robust to changes in window sizes.
- Due to lower chances of signal interference from adjacent rebar, the elements with sparser distributed rebar are more recognizable by GPR scanners.

In this research, AlexNet has been only employed to detect the existence of rebar. As part of future research directions, the authors plan to focus on recognizing size and depth of rebar with AlexNet. The authors plan to handle this task by considering the relations between the spatial information of the rebar and the corresponding coordinates in GPR imagery data.